\documentclass[runningheads]{llncs} 

\usepackage{amsfonts,amsmath}
\usepackage{easyeqn}
\usepackage{multirow}
\usepackage{subfigure}
\usepackage{graphicx}
\usepackage{booktabs}
\usepackage{numprint}

\def\RR{\mathbb{R}}
\def\EE{\mathbb{E}}
\def\xv{\mathbf{x}}
\def\yv{\mathbf{y}}
\def\uv{\mathbf{u}}
\def\vv{\mathbf{v}}

\def\ev{\mathbf{e}}
\def\t{\mathrm{\times}}
\newcommand{\T}{\mathrm{T}}

\def\ZV{\mathbf{Z}}

\def\ZC{\mathcal{Z}}
\def\FC{\mathcal{F}}
\def\PC{\mathcal{P}}
\def\RC{\mathcal{R}}
\def\LC{\mathcal{L}}
\def\CC{\mathcal{C}}
\def\eps{\epsilon}

\def\observedIndices{\Omega}
\def\targetMatrix{M}
\def\parameterMatrix{W}
\def\xFactor{U}
\def\yFactor{V}
\def\penaltyFunction{R}
\def\nsize{n}
\def\msize{m}
\def\zSet{\ZC}

\def\lipshitzConst{L}

\DeclareMathOperator{\vect}{vec}

\begin{document}
\title{Sparse Group Inductive Matrix Completion}

\author{Ivan Nazarov\inst{1,2}\orcidID{0000-0003-2890-3095} \and
Boris Shirokikh\inst{1,2,3} \and
Maria Burkina\inst{2,3} \and
Gennady Fedonin\inst{2,4} \and
Maxim Panov\inst{1,2}\orcidID{0000-0001-5161-2822}}

\authorrunning{I. Nazarov et al.}

\institute{Skolkovo Institute of Science and Technology (Skoltech), Moscow, Russia \and
Institute for Information Transmission Problems of RAS, Moscow, Russia \\
 \and
Moscow Institute of Physics and Technology, Dolgoprudny, Russia \\
 \and
Central Research Institute for Epidemiology, Moscow, Russia \\
\email{m.panov@skoltech.ru}, \email{ivan.nazarov@skolkovotech.ru}}

\maketitle              

\begin{abstract}
  We consider the problem of matrix completion with side information (\textit{inductive matrix completion}). In real-world applications many side-channel features are typically non-informative making feature selection an important part of the problem. We incorporate feature selection into inductive matrix completion by proposing a matrix factorization framework with group-lasso regularization on side feature parameter matrices. We demonstrate, that the theoretical sample complexity for the proposed method is much lower compared to its competitors in sparse problems, and propose an efficient optimization algorithm for the resulting low-rank matrix completion problem with sparsifying regularizers. Experiments on synthetic and real-world datasets show that the proposed approach outperforms other methods. 
\end{abstract}

\section{Introduction}
\label{sec:intro}
  Matrix completion methods are widely used in the variety of applications such as recommender systems~\cite{Rennie2005,Koren2009}, clustering~\cite{Yi2012}, multi-label learning~\cite{Argyriou2008,Cabral2011}, signal processing~\cite{Weng2012}, computer vision~\cite{Chen2004}, etc. The traditional low-rank matrix completion acts in the transductive setting, in which matrix elements are reconstructed as an inner product of factors (sometimes even implicitly). Each factor corresponds either to a row variable or to the column one and can be interpreted as a latent representation, which is learned for each variable based on the given subset of matrix elements. Such an approach is supported by a well-developed theory~\cite{Candes2009,Candes2010}, which in particular states that to recover a \(\nsize_1 \t \nsize_2\) low-rank matrix it suffices to know \(O(N \log^2 N)\) entries sampled uniformly at random where \(N = \max\{\nsize_1, \nsize_2\}\). We also note that a distribution-free bound requires \(O(N^{3/2})\) entries for exact recovery~\cite{Shamir2014}.
  
  However, often in addition to the partially observed matrix there exists information about row- and column- objects. For example, \textit{side-channel information} in the form of user profiles and movie genres is natural in recommender systems and was proved to be useful in real-world applications~\cite{Hannon2010}. Such information is especially important in a cold-start scenario, if no interaction has been recorded so far between new object and the already existing ones, and thus the prediction is possible only if based on side information. Many approaches proposed recently are focused on direct incorporating side features into matrix completion algorithms under the name of \textit{inductive matrix completion} (IMC); see~\cite{Xu2013,Jain2013,Natarajan2014,Chiang2015,Si2016,Lu2016,Guo2017,Zhang2018} among many other contributions. On a theoretical side, the main result, that have been obtained so far, is that the sample complexity can be reduced to \(O(\log N)\) if the features have a good predictive power. We note that it is natural to expect that not all the features are predictive and it is desirable to design algorithms which would be robust to features without any predictive power. Surprisingly, only few attempts to introduce sparsity into IMC were made~\cite{Lu2016,Soni2017} and many questions still remain to be unsolved, which leaves vast room for further improvements in the area.
  
  In this paper we introduce a new matrix recovery algorithm that allows to efficiently filter out non-informative side-channel features. Our algorithm is based on a matrix factorization approach and achieves sparsity by a sparse-group penalty on factor matrices, at that the groups correspond to the columns of factor matrices. We propose an optimization algorithm for the resulting non-convex non-(bi)quadratic problem, which is based on the ideas of ADMM~\cite{bertsekas1989parallel,boydetal2011}. Our approach has proven theoretical guarantees, which allow for improved generalization bounds in the sparse case, i.e. when many side-channel features are not predictive. In addition, we provide empirical evidence that the proposed approach efficiently functions on a variety of synthetic and real-world datasets. The state-of-the-art performance is demonstrated in solving the problems exhibiting quite a lot of uninformative features.
  Our contribution can be summarized in the following way:
  \begin{itemize}
    \item We introduce new matrix completion algorithm based on the ideas of matrix factorization and the usage of a carefully selected sparsity inducing penalty. This algorithm is able to filter out uninformative side-channel features.
    
    \item We study the problem theoretically and prove that the proposed approach is effective in feature selection. In particular, the sample complexity of the obtained algorithm is lower than in the case with the one without feature selection if applied to data with uninformative features.
    
    \item An efficient optimization scheme proposed in our algorithm allows to effectively solve practical large-scale problems. A Python implementation of the algorithm is openly available\footnotemark \footnotetext{https://github.com/premolab/SGIMC}.
  \end{itemize}
  We begin the paper with a brief overview of the related literature in Section~\ref{sec:related_work}, then we introduce the matrix completion optimization problem with the sparse-group penalty in Section~\ref{sec:problem_statement}, and outline the optimization algorithm in Section~\ref{sec:the_algorithm}. In Section~\ref{sec:theory} we derive the generalization error bounds for the proposed model, and conduct numerical experiments in Section~\ref{sec:experiments}. The research is concluded by some remarks and the outlook in Section~\ref{sec:conclusion}.

\subsection{Related work}
\label{sec:related_work}
  Extensive study of inductive matrix completion problems was initiated by the paper~\cite{Xu2013} showing that \(O(\log N)\) samples are sufficient for the exact matrix recovery performed by an algorithm with a nuclear norm penalty. The matrix factorization approach based on alternating minimization was proposed in~\cite{Jain2013}, while recently it has been amended with a new proximal optimization algorithm~\cite{Zhang2018} that achieves the state-of-the-art computational and sample complexity. The paper~\cite{Chiang2015} considers the situation with the imperfect features that are not informative enough to fully reconstruct the target matrix. They combine inductive and transductive matrix completion approaches to achieve a consistent recovery in case of both perfect and imperfect features. The paper~\cite{Lu2016} adds sparsity penalty to inductive matrix completion, then analyzes the performance of the resulting algorithm, and proves relevant sample complexity bounds. This approach has a high computational and memory complexity at it requires the computation of the nuclear norm for the matrix of the same size as the one being recovered. Moreover, it provides no analysis of the impact of the feature selection on the solution. The combinatorial optimization algorithm that achieves fast recovery rates in the noisy case, is proposed in~\cite{Soni2017}. In the related setting of the inductive link prediction in graphs~\cite{Baldin2018} the minimax optimal rates in sparse and low-rank regime are obtained and the tradeoffs between the statistical rates and the computational complexity are uncovered.

\section{Sparsifying the inductive matrix completion}
\label{sec:problem_statement}
  Let \(\targetMatrix \in \RR^{\nsize_1 \times \nsize_2}\) be the target matrix, and suppose that only elements \(M_{ij}\) for \((i, j) \in \observedIndices \subset \{1, \dots, \nsize_1\} \times \{1, \dots, \nsize_2\}\) are observed. Suppose that the side-channel information is observed as well and is summarized in the matrices \(X \in \RR^{\nsize_1 \times d_1}\) and \(Y \in \RR^{\nsize_2 \times d_2}\) for the rows and columns of \(\targetMatrix\), respectively. Furthermore, assume that side information is predictive of the entries in \(\targetMatrix\) through a bilinear model, i.e. \(\targetMatrix_{ij} \sim \xv_i^{\T} \parameterMatrix \yv_j\) for some matrix \(\parameterMatrix \in \RR^{d_1 \times d_2}\). The goal of the inductive matrix completion problem is to recover missing entries in \(\targetMatrix\) from \(M_{\observedIndices}\) as well as the side-channel information \(X\) and \(Y\).
  
  The majority of the existing inductive matrix completion methods~\cite{Jain2013,Xu2013,Chiang2015} assume that matrix \(\parameterMatrix\) has rank \(k < \min(d_1, d_2)\). There are two main approaches to enforce a low rank of matrix \(\parameterMatrix\):
  \begin{enumerate}
    \item To penalize \(\parameterMatrix\) with nuclear norm penalty \(\|\parameterMatrix\|_*\), which would result in low-rank solutions of the matrix completion optimization problem~\cite{Xu2013,Chiang2015,Lu2016}.
    \item To parametrize \(\parameterMatrix\) by \(\xFactor \yFactor^{\T}\) for some matrices \(\xFactor \in \RR^{ d_1 \times k}\) and \(\yFactor \in \RR^{d_2 \times k}\), see~\cite{Jain2013,Zhang2018}.
  \end{enumerate}
  In addition to modelling low-rank structure explicitly, the latter approach reduces the computational complexity as compared to dealing with a full matrix \(\parameterMatrix\). We proceed with the parametrization \(\parameterMatrix = \xFactor \yFactor^{\T}\), since our aim is to solve high-dimensional practical problems.
  
  Suppose \(\LC\colon \RR \times \RR \to \RR\) is the loss function (convex) and consider the problem:
  \begin{EQA}[c]
  \label{imc_model}
    \min_{\xFactor, \yFactor} \sum_{(i, j) \in \observedIndices} \LC\bigl(\targetMatrix_{ij}, (X \xFactor \yFactor^{\T} Y^{\T})_{ij}\bigr)
    +
    \lambda_{\xFactor} \penaltyFunction(\xFactor) + \lambda_{\yFactor} \penaltyFunction(\yFactor)
    \,,
  \end{EQA}
  where \(\penaltyFunction(\cdot)\) is the penalty function and \(\lambda_{\xFactor}, \lambda_{\yFactor} \ge 0\) are penalty coefficients. The usual choice~\cite{Chiang2015} of \(\penaltyFunction(\cdot)\) is the squared Frobenius norm of the argument matrix \(Z \in \RR^{d \t k}\): \(\penaltyFunction(Z) = \|Z\|_F^2 = \sum_{i = 1}^{d} \sum_{j = 1}^{k} z_{ij}^2\), which is equivalent to the Tikhonov regularization in a classical linear regression.

  In this work we propose to use the sparsity inducing penalty in the form of the matrix group norm: \(\penaltyFunction(Z) = \|Z\|_{2, 1} = \sum_{i = 1}^{d} \|\ev_{i}^{\T} Z\|_{2}\), where by \(\ev_i\) we denote a unit column vector with \(1\) at the \(i\)-th place, and zeros everywhere else. Thus, the optimization problem of the IMC with sparse-group penalty (SGIMC) is as follows:
  \begin{EQA}[c]
  \label{sparse_imc_model}
    \min_{\xFactor, \yFactor} \sum_{(i, j) \in \observedIndices} \LC\bigl(\targetMatrix_{ij}, (X \xFactor \yFactor^{\T} Y^{\T})_{ij}\bigr)
    +
    \lambda_{\xFactor} \|\xFactor\|_{2, 1}
    +
    \lambda_{\yFactor} \|\yFactor\|_{2, 1}
    \,.
  \end{EQA}
  This composite penalty encourages the group sparsity over the rows of \(\xFactor\) and \(\yFactor\), which effectively performs a feature selection in side-channel information \(X\) and \(Y\), respectively.
  
  We note that the basis of the algorithmic approach proposed in this paper allows us to deal with a broad choice of penalty functions and combinations thereof. In particular, we also consider the squared Frobenius norm \(\penaltyFunction(Z) = \|Z\|_{\rm F}^2\) and the matrix \(L_1\)-norm \(\penaltyFunction(Z) = \|Z\|_{1, 1} = \sum_{i = 1}^{\nsize} \sum_{j = 1}^{d} |z_{ij}|\).
  
  The solutions for~\eqref{imc_model} and~\eqref{sparse_imc_model} may achieve an accurate matrix completion only if the features \(\xv_i\) and \(\yv_j\) have a good predictive power. For example, the necessary condition for the possibility of the exact recovery, i.e. \(M = X W Y^\T\), is
  \begin{EQA}[c]
  \label{perfect_condition}
    \mathrm{col}(\targetMatrix) \subset \mathrm{col}(X), \quad \mathrm{row}(\targetMatrix) \subset \mathrm{col}(Y)
    \,,
  \end{EQA}
  where \(\mathrm{col}(\cdot)\) and \(\mathrm{row}(\cdot)\) denote the column- and row-spaces, respectively.
  Such a side-channel feature set can be considered ``perfect'' since it fully describes the structure of matrix \(\targetMatrix\) via the IMC model. In practice, however, the identity \(\targetMatrix = X \parameterMatrix Y^{\T}\) may fail to hold for any choice of \(\parameterMatrix\).
  If the condition~\eqref{perfect_condition} is violated, while matrix \(\targetMatrix\) is still low-rank, then the recovery is still possible through classical, transductive matrix completion methods~\cite{Candes2009}. It is straightforward to combine the inductive and transductive matrix completion, e.g. by considering the following completion model
  \begin{EQA}
    && \min_{\xFactor, \yFactor, \tilde{\xFactor}, \tilde{\yFactor}} \sum_{(i, j) \in \observedIndices} \LC\bigl(\targetMatrix_{ij},  (X \xFactor \yFactor^{\T} Y^{\T} + \tilde{\xFactor} \tilde{\yFactor}^{\T})_{ij}\bigr)
    \\
    && \quad +
    \lambda_{\xFactor} \|\xFactor\|_{2, 1}
    +
    \lambda_{\yFactor} \|\yFactor\|_{2, 1}
    +
    \lambda_{\tilde{\xFactor}} \|\tilde{\xFactor}\|_{F}^2
    +
    \lambda_{\tilde{\yFactor}} \|\tilde{\yFactor}\|_{F}^2
    \,,
  \end{EQA}
  where \(\tilde{\xFactor} \in \RR^{\nsize_1 \t k_1}, \tilde{\yFactor} \in \RR^{\nsize_2 \t k_1}\) for some \(k_1 < \max(\nsize_1, \nsize_2)\).
  The similar approach though without the sparsity regularization is considered in~\cite{Chiang2015}, in which the nuclear norm minimization approach is used for the solution. A different approach based on nuclear norm minimization, though with an additional sparsity constraint, is considered in~\cite{Lu2016}. We note that our approach seems to be more preferable from the computational point of view as it does not deal directly with \(\nsize_1 \t \nsize_2\) matrices and does not require an expensive computation of the nuclear norm.

\section{Algorithm} 
\label{sec:the_algorithm}
  In this section we provide further details on the numerical procedure for solving SGIMC problem. We begin with the problem statement, then move on to the algorithm and the derivation thereof and conclude with a brief discussion.

\subsection{The optimization problem and the algorithm} 
\label{sub:the_optimization_problem_and_the_algorithm}
  SGIMC problem for the dataset \((M_{\observedIndices}, X, Y)\) can be formulated as the following optimization problem. For a given rank \(k \geq 1\) find \(\xFactor \in \RR^{d_1 \times k}\) and \(\yFactor \in \RR^{d_2 \t k}\) that minimize
  \begin{equation}
  \label{eq:imc_problem}
    \begin{aligned}
      J(\xFactor, \yFactor)
      = \sum_{\omega \in \observedIndices}
        \LC(M_\omega, p_\omega)
        + \lambda_{\xFactor} \|\xFactor\|_{2,1}
        + \lambda_{\yFactor} \|\yFactor\|_{2,1}
      \,,
    \end{aligned}
  \end{equation}
  where \(p_\omega = \ev_i^\T X \xFactor \yFactor^\T Y^\T \ev_j \,, \omega = (i, j)\) and \(\LC(y, p)\) is the loss function. Loss \(\LC\) is determined by a particular matrix reconstruction problem. For example, in the regression case \(M\) is real-valued and the usual choice of loss is \(\LC(y, p) = \tfrac{1}{2} (y - p)^2\), i.e. \(L_2\) reconstruction loss. In the classification case \(M\) is \(\pm 1\)-valued and the standard choice is the log-loss \(\LC(y, p) = \log\bigl(1 + e^{-y p}\bigr)\).
  %
  The sparse-group penalty performs a feature selection by zeroing entire rows in each matrix, because the reconstruction is based on the terms like \(X \xFactor = \sum_{p=1}^{d_1} (X \ev_p) (\ev_p^\T \xFactor)\) and \(X \ev_p\) is the \(p\)-th feature of \(X\). We note that the optimization algorithm described below allows for the broad selection of penalty functions, including the squared Frobenius norm and matrix \(L_1\) norm, which are also implemented in our Python package.

  The problem~\eqref{eq:imc_problem} is bi-convex, i.e. it is convex with respect to one matrix while the other is fixed, e.g. \(\xFactor \mapsto J(\xFactor, \yFactor)\) is convex for fixed \(\yFactor\) and \textit{vice versa}. A natural optimization approach is the nonlinear Gauss-Seidel iterations, i.e. coordinatewise iterative descent, which alternates between
  \begin{equation} \label{eq:coordinatewise_descent}
    \begin{aligned}
      \xFactor_{t + 1}
      & =
      \arg \min_{\xFactor \in \RR^{d_1 \times k}} J(\xFactor, \yFactor_{t})
      \,, \\
      \yFactor_{t + 1}
      & = 
      \arg \min_{\yFactor \in \RR^{d_2 \times k}} J(\xFactor_{t + 1}, \yFactor)
      \,,
    \end{aligned}
  \end{equation}
  until a convergence criterion is met, e.g. \(\xFactor_t \yFactor_t^\T\) changed insignificantly. The iterations converge, since each subproblem has a unique solution guaranteed by the strong convexity of the penalty~\cite{bertsekas1989parallel}.

\subsection{The partial problem} 
\label{sub:the_partial_problem}
  In this subsection we show that the iterations~\eqref{eq:coordinatewise_descent} require considering of only one subproblem of~\eqref{eq:imc_problem}, and propose a procedure to solve it.

  The structure of the loss term and the regularizer in~\eqref{eq:imc_problem} imply that the objective \(J\) for \((M_{\observedIndices}, X, Y)\) is the same as the objective function \(J^\T\) for the parameters \((M_{\observedIndices}^\T, Y, X)\) and with the arguments \(\xFactor\) and \(\yFactor\) swapped, i.e. the ``transposed'' problem. Therefore the partial objective \(\yFactor \to J(\xFactor, \yFactor)\) for a fixed \(\xFactor\) is the same as the partial objective \(\xFactor \to J^\T(\yFactor, \xFactor)\) for the transposed problem, which implies that it suffices to propose an algorithm for solving \(\min_{\xFactor} J(\xFactor, \yFactor)\) for \((M_{\observedIndices}, X, Y)\) and fixed \(\yFactor\) to be able to run~\eqref{eq:coordinatewise_descent}.

  After dropping the constant terms from the partial problem~\eqref{eq:imc_problem} for parameters \((M_{\observedIndices}, X, Y)\) with respect to \(\xFactor\) while holding \(\yFactor\) fixed, we get the partial problem in the following form:
  \begin{equation} \label{eq:imc_problem_partial}
    \begin{aligned}
      & \underset{\xFactor \in \RR^{d_1\times k}}{\min}
      & & \sum_{\omega \in \observedIndices}
      \LC(M_\omega, p_\omega) + \lambda_{\xFactor} R(\xFactor) \,,
    \end{aligned}
  \end{equation}
  where \(p_\omega = \ev_i^\T \bigl(X \xFactor Q^\T\bigr) \ev_j\) for \(\omega = (i, j)\) and \(Q = Y \yFactor\) is an \(n_2 \times k\) fixed matrix.

  We propose to solve~\eqref{eq:imc_problem_partial} with the Alternating Direction Method of Multipliers (ADMM), that was initially proposed in~\cite{glowinskimarroco1975} and~\cite{gabaymercier1976}, while its convergence guarantees were derived in~\cite{gabay1983} and~\cite{ecksteinbertsekas1992}. Applied to~\eqref{eq:imc_problem_partial}, the ADMM iterations take the following form:
  \begin{align}
    \xFactor_{t + 1}
    &= \arg \min_{\xFactor}
      \sum_{\omega \in \observedIndices} \LC(M_\omega, p_\omega)
      + \tfrac{\lambda_\mathrm{R}}{2} \|\xFactor\|_F^2
    + \tfrac{1}{2\eta}\|\xFactor - (Z_t - \Phi_t) \|_F^2
    \,, \label{eq:imc_prtl_admm_x} \\
    Z_{t+1}
    &= \arg \min_Z ~
      \lambda_{\xFactor} \|Z\|_{2, 1}
      + \tfrac{1}{2\eta}\|Z - (\xFactor_{t+1} + \Phi_t)\|_F^2
    \,, \label{eq:imc_prtl_admm_z} \\
    \Phi_{t+1}
    &= \Phi_t + (\xFactor_{t+1} - Z_{t+1})
    \,, 
  \end{align}
  where \(\eta > 0\), the dual variable \(\Phi\) is a \(d_1 \times k\) matrix and \(\tfrac{\lambda_\mathrm{R}}{2} \|\xFactor\|_F^2\) is an additional penalization term ensuring the strong convexity of the objective on the corresponding step of the algorithm.

\subsubsection{Solving the \(u\)-step}
\label{ssub:solving_eq_imc_prtl_admm_x}
  The \(u\)-step,~\eqref{eq:imc_prtl_admm_x}, is a differentiable convex minimization problem with quadratic regularization, which admits an exact solution for \(L_2\) loss. When the loss \(\LC\) is a more general loss, e.g. the log-loss, this subproblem is solved numerically. A similar problem without the ADMM-based quadratic term is solved in~\cite{yuetal2014} using the trust region conjugate gradient method (TRON) proposed in~\cite{linetal2008} for large-scale linear models. The gradient and the ``hessian-vector'' product for~\eqref{eq:imc_prtl_admm_x} at some \(\xFactor\), needed by conjugate methods, are
  \begin{align}
    {\tt grad}_{\xFactor}
    &= X^\T G Q + \bigl(\lambda_\mathrm{R} + \tfrac1\eta\bigr) \xFactor
    - \tfrac{1}{\eta} \bigl(Z_t - \Phi_t\bigr)
    \,, \label{eq:prtl_admm_x_grad} \\
    {\tt HessV}_{\xFactor}(D)
    &= X^\T \bigl(H \odot (X D Q^\T)\bigr) Q
    + \bigl(\lambda_\mathrm{R} + \tfrac1\eta\bigr) D
    \,, \label{eq:prtl_admm_x_hessv}
  \end{align}
  respectively. Here \(X\) and \(Q\) are as in~\eqref{eq:imc_problem_partial}, \(D \in \RR^{d_1 \times k}\) is the ``vector'' for the hessian-vector product, \(\odot\) is the Hadamard product of two conforming matrices, and \(G\) and \(H\) are \(n_1 \times n_2\) matrices with the sparsity structure of \(M\) and elements \(\LC'(M_\omega, p_\omega)\) and \(\LC''(M_\omega, p_\omega)\), respectively. 
  With \(\vect(\cdot)\) denoting the row-major vectorization, these expressions follow from the chain rule and the \(\vect{A B C} = (A \otimes C^\T) \vect{B}\) identity, which is implied by the structure of the Kronecker product of \(A\) and \(C^\T\) and the row-major vectorization~\cite{cookbook2012}.

  The paper~\cite{yuetal2014} derives the fast basic matrix operations that are required for computing \eqref{eq:prtl_admm_x_grad} and \eqref{eq:prtl_admm_x_hessv}. The \(S \to X^\T S Q\) maps \(\observedIndices\)-sparse \(\nsize_1 \t \nsize_2\) matrix \(S\) to dense \(d_1 \times k\) matrix, and \(D \to X D Q^\T\) maps a dense \(d_1 \times k\) matrix \(D\) to an \(\observedIndices\)-sparse \(\nsize_1 \t \nsize_2\) matrix with the same sparsity as \(M\). For a dense matrix \(Q\) both operations have overall complexity \(O\bigl((\lvert\observedIndices\lvert + {\tt nnz}(X))\cdot k\bigr)\), where \({\tt nnz}(X)\) is \(n_1 d_1\) if \(X\) is dense, and the number of nonzero elements if \(X\) is sparse.


\subsubsection{Solving the \(z\)-step}
\label{ssub:solving_eq_imc_prtl_admm_z}
  The objective in step~\eqref{eq:imc_prtl_admm_z} is additively separable across the  rows of matrix \(Z\), and therefore decomposes into \(d_1\) independent subproblems:
  \begin{equation*} \label{eq:admm_z_subprb}
    z_j = \arg \min_z
    \tfrac{1}{2} \|z - a_j\|_2^2
      + \eta \lambda_{\xFactor} \|z\|_2
      \,,\, j = 1\ldots d_1
    \,,
  \end{equation*}
  where \(a_j = e_j^\T(\xFactor_{t+1} + \Phi_t)\). This problem admits an efficiently computable solution~\cite{simonetal2013}:
  \begin{equation*}
  \label{eq:sparse_group_shrinkage}
    z_j = S_\mathrm{G}\bigl(a_j; \eta \lambda_{\xFactor}\bigr)
    \,,
  \end{equation*}
  where \(S_\mathrm{G}(u; \nu) = \bigl(1 - \tfrac{\nu}{\|u\|_2}\bigr)_+ u\) is the group shrinkage and \((x)_+ = \max\{x, 0\}\).







\section{Theoretical analysis}
\label{sec:theory}
  In this section we analyze the impact of the feature selection on the prediction quality of matrix completion methods. We consider the case of ``perfect'' and ``imperfect'' side-channel features separately, and demonstrate that in the presence of redundant features the sparse-group penalty achieves lower sample complexity compared to the Frobenius norm penalty.

\subsection{Perfect features}
  Our first result deals with ``perfect'' features in terms of the conditions~\eqref{perfect_condition}. Below we will assume that loss function \(\LC\) is Lipshitz with Lipshitz constant \(\lipshitzConst\). The matrix \(\targetMatrix\) is assumed to be partially observed with indices \((i, j) \in \observedIndices\) drawn independently from some distribution on index pairs.
  We consider the following subproblem first:
  \begin{EQA}[c]
  \label{imc_perfect}
    \min_{\xFactor, \yFactor} \sum_{i = 1}^{\nsize_1} \sum_{j = 1}^{\nsize_2} \LC\bigl(\targetMatrix_{ij}, (X \xFactor \yFactor^{\T} Y^{\T})_{ij}\bigr)
    \,,
  \end{EQA}
  with \(\xFactor \in \RR^{d_1 \t k}, \yFactor \in \RR^{d_2 \t k}\) for \(k = \mathrm{rank}(\targetMatrix)\), and denote its solution as \(\hat{\xFactor}, \hat{\yFactor}\). In the realizable case the solution gives zero value to the objective in~\eqref{imc_perfect} leading to strict equality \(M = X \hat{\xFactor} \hat{\yFactor}^{\T} Y^{\T}\).
  
  If some of the features are not predictive then matrices \(\hat{\xFactor}\) and \(\hat{\yFactor}\) are row-sparse. Therefore, if \(\hat{\xFactor}\) and \(\hat{\yFactor}\) have no more than \(s_1\) and \(s_2\) non-zero rows respectively, we can bound
  \begin{EQA}[c]
    \|\hat{\xFactor}\|_{2, 1} \leq s_1 \sqrt{k} ~ u_{\infty}
    \,,
    \quad
    \|\hat{\yFactor}\|_{2, 1} \leq s_2 \sqrt{k} ~ v_{\infty}
    \,,
  \end{EQA}
  where \(u_{\infty}\) and \(v_{\infty}\) are the maximal values in matrices \(\hat{\xFactor}\) and \(\hat{\yFactor}\), respectively. We further assume that feature vectors are normalized so that
  \begin{EQA}[c]
    \|\xv_i\|_{\infty} \le 1, \quad \|\yv_j\|_{\infty} \le 1
    \,.
  \end{EQA}
  
  We denote that \(\xFactor_{2, 1} = \|\hat{\xFactor}\|_{2, 1}, \yFactor_{2, 1} = \|\hat{\yFactor}\|_{2, 1}\) and consider the following constraints on the parameters of the model: \(\|\xFactor\|_{2, 1} \le \xFactor_{2, 1}, \|\yFactor\|_{2, 1} \le \yFactor_{2, 1}\). Let \(\msize\) be the size of the known element set \(\observedIndices\).
  A suitable choice of \(\lambda_{\xFactor}\) and \(\lambda_{\yFactor}\) makes the problem~\eqref{sparse_imc_model} equivalent to the following problem:
  \begin{EQA}
  \label{sparse_imc_model_theory}
    && \min_{\xFactor, \yFactor} \sum_{(i, j) \in \observedIndices} \LC(\targetMatrix_{ij},  (X \xFactor \yFactor^{\T} Y^{\T})_{ij}),
    \\
    && \text{subject to} ~~
    \|\xFactor\|_{2, 1} \le \xFactor_{2, 1}
    , ~
    \|\yFactor\|_{2, 1} \le \yFactor_{2, 1}
    \,.
  \end{EQA}
  %
  We introduce the empirical risk and the expected risk respectively:
  \begin{EQA}
    \hat{r}_{\msize}(\xFactor, \yFactor)
    &=& \frac{1}{\msize} \sum_{(i, j) \in \observedIndices} 
      \LC(\targetMatrix_{ij}, \xv_{i}^{\T} \xFactor \yFactor^{\T} \yv_{j})
    \,, \\
    r_{\msize}(\xFactor, \yFactor)
    &=& \EE_{(i, j)} \LC(\targetMatrix_{ij}, \xv_{i}^{\T} \xFactor \yFactor^{\T} \yv_{j})
    \,.
  \end{EQA}
  Under the constraints in~\eqref{sparse_imc_model_theory} a solution with zero empirical error \(\hat{r}_m(\xFactor, \yFactor)\) is feasible, which implies that bounding the generalization error, \(r_m(\xFactor, \yFactor)\), amounts to bounding just the empirical Rademacher complexity \(\widehat{\RC}_{\msize}(\FC)\), see~\cite{Bartlett2002,Maurer2012}.
  %

  The obtained results can be summarized by the following theorem, which proof can be found in Appendix~\ref{sec:proofTheorem}.
  \begin{theorem}
  \label{theorem:main}
    Consider problem~\eqref{sparse_imc_model_theory} with \(\xFactor_{2, 1} = \|\hat{\xFactor}\|_{2, 1}, ~ \yFactor_{2, 1} = \|\hat{\yFactor}\|_{2, 1}\) and  \(\hat{\xFactor} \in \RR^{d_2 \t k}, \hat{\yFactor} \in \RR^{d \t k}\) being the solutions of the problem~\eqref{imc_perfect}. Then the expected risk of an optimal solution \((\xFactor^*, \yFactor^*)\) with probability at least \(1 - \delta\) satisfies the following inequality:
    \begin{EQA}[c]
      r_{\msize}(\xFactor^*, \yFactor^*)
      \le
      \xFactor_{2, 1} \yFactor_{2, 1} \lipshitzConst \frac{2^{3/2}}{\sqrt{\msize}} \Bigl(2 + \sqrt{\ln (d_1 d_2)}\Bigr)
      +
      \sqrt{\frac{9 \ln 2 / \delta}{2 \msize}}
      \,.
    \end{EQA}
    Moreover, if the ideal solution is sparse, i.e. \(\hat{\xFactor}\) and \(\hat{\yFactor}\) have no more than \(s_1\) and \(s_2\) non-zero rows, respectively, with probability at least \(1 - \delta\) it holds
    \begin{EQA}[c]
      r_{\msize}(\xFactor^*, \yFactor^*)
      \le
      s_1 s_2 k u_{\infty} v_{\infty} \lipshitzConst \frac{2^{3/2}}{\sqrt{\msize}} \Bigl(2 + \sqrt{\ln (d_1 d_2)}\Bigr)
      +
      \sqrt{\frac{9 \ln 2 / \delta}{2 \msize}}
      \,,
    \end{EQA}
    where \(u_{\infty}, v_{\infty}\) are the maximal values in matrices \(\hat{\xFactor}\) and \(\hat{\yFactor}\), respectively.
  
    Finally, if feature vectors \(\xv_i\) and \(\yv_j\) are sparse, then the bound can be improved: with probability at least \(1 - \delta\) it holds
    \begin{EQA}[c]
      r_{\msize}(\xFactor^*, \yFactor^*)
      \le
      s_1 s_2 k u_{\infty} v_{\infty} \lipshitzConst \frac{2^{3/2}}{\sqrt{\msize}} \left(2 + \sqrt{\ln (r_1 r_2)}\right)
      +
      \frac{2 \lipshitzConst}{\sqrt{\msize}}
      +
      \sqrt{\frac{9 \ln 2 / \delta}{2 \msize}}
      \,,
    \end{EQA}
    where \(r_1\) and \(r_2\) are such that all \(\xv_i\) and \(\yv_j\) have no more than \(r_1\) and \(r_2\) non-zero coordinates respectively.
  \end{theorem}
  We note that the use of the Frobenius norm penalization allows us to bound the sample complexity of \(\epsilon\)-recovery in the case of sparse coefficient matrices by \(O(s_1 s_2 k^2 d_1 d_2 \ln (d_1 d_2) / \epsilon^2)\), while for our method it can be estimated as \(O(s_1^2 s_2^2 k^2 \ln (d_1 d_2) / \epsilon^2)\), which leads to a faster recovery in large-\(d\) small-\(s\) setting. In the case when features vectors are also sparse, the Frobenius norm penalization gives \(O(s_1 s_2 k^2 r_1 r_2 \ln (r_1 r_2) / \epsilon^2)\), while for our method the bound is \(O(s_1^2 s_2^2 k^2 \ln (r_1 r_2) / \epsilon^2)\), so that the performance differences depend on the relative values of \(s_1, s_2, r_1\) and \(r_2\).
  
\subsection{Imperfect features}
  In the real-world situations features are usually far from being ``perfect'', so the measuring the feature quality becomes important. Since the bias of the best possible estimator~\eqref{imc_perfect} can be expressed as \(\hat{N} = \targetMatrix - X \hat{\xFactor} \hat{\yFactor}^{\T} Y^{\T}\), we use the Frobenius norm of residuals matrix \(N_F = \|\hat{N}\|_{F}\) to measure the feature quality. We consider the following problem
  \begin{EQA}
    \label{sparse_imc_model_theory_dirty}
    && \min_{\xFactor, \yFactor, \tilde{\xFactor}, \tilde{\yFactor}} \sum_{(i, j) \in \observedIndices} \LC(\targetMatrix_{ij},  (X \xFactor \yFactor^{\T} Y^{\T} + \tilde{\xFactor} \tilde{\yFactor}^{\T})_{ij}),
    \\
    && \text{subject to} ~~
    \|\xFactor\|_{2, 1} \le \xFactor_{2, 1}
    , ~
    \|\yFactor\|_{2, 1} \le \yFactor_{2, 1},
    \\
    && \qquad \qquad ~~ \frac{1}{2}(\|\tilde{\xFactor}\|_{F}^{2} + \|\tilde{\yFactor}\|_{F}^{2}) \le N_F
    \,,
  \end{EQA}
  where \(\tilde{\xFactor} \in \RR^{\nsize_1 \t k_1}, \tilde{\yFactor} \in \RR^{\nsize_2 \t k_1}\) and \(k_1\) is the rank of matrix \(\hat{N}\).
  
  We note that the condition \(\frac{1}{2}(\|\tilde{\xFactor}\|_{F}^{2} + \|\tilde{\yFactor}\|_{F}^{2}) \le N_F\) effectively bounds the nuclear norm of \(\tilde{\xFactor} \tilde{\yFactor}^{\T}\) by the nuclear norm of \(\hat{N}\). This follows from the facts that \(\|\tilde{\xFactor} \tilde{\yFactor}^{\T}\|_* \le \frac{1}{2}(\|\tilde{\xFactor}\|_{F}^{2} + \|\tilde{\yFactor}\|_{F}^{2})\) and \(\|\hat{N}\|_{F} \le \|\hat{N}\|_{*}\). The result for the imperfect features case follows from Theorem~1 of~\cite{Chiang2015}:
  \begin{theorem}
  \label{noisy_bound}
    Consider problem~\eqref{sparse_imc_model_theory_dirty} with \(\xFactor_{2, 1} = \|\hat{\xFactor}\|_{2, 1}, ~ \yFactor_{2, 1} = \|\hat{\yFactor}\|_{2, 1}\) and  \(\hat{\xFactor} \in \RR^{d_2 \t k}, \hat{\yFactor} \in \RR^{d \t k}\) being the solutions of the problem~\eqref{imc_perfect}. Then with probability at least \(1 - \delta\) the expected risk \(r_m(f^*)\) of an optimal solution \(f^* = (\xFactor^*, \yFactor^*, \tilde{\xFactor}^*, \tilde{\yFactor}^*)\) is upper bounded by
    \begin{EQA}
      && r_{\msize}(f^*)
      \le
      C \lipshitzConst \xFactor_{2, 1} \yFactor_{2, 1} \sqrt{\frac{\ln (d_1 d_2)}{\msize}}
      +
      \sqrt{\frac{9 \ln 2 / \delta}{2 \msize}}
      \\
      &+&
      \min \Bigl\{4 \lipshitzConst N_F  \sqrt{\frac{\log 2 \nsize}{\msize}}, \sqrt{C \lipshitzConst \frac{N_F (\sqrt{n_1} + \sqrt{n_2})}{\msize}}\Bigr\}
      \,,
    \end{EQA}
    where \(C\) is some absolute constant.
  \end{theorem}
  We note that this result can be further specified to be adapted to particular sparsity patterns in the considered parameter matrices and feature vectors. The bound of Theorem~\ref{noisy_bound} still enjoys the benefits of sparsity, while achieving sample complexity of \(\epsilon\)-recovery of \(O(N^{3/2})\) in the worst case for \(N = \max\{n_1, n_2\}\). If the features are predictive, i.e. \(N_F = o(N)\), then the sample complexity improves to \(o(N^{3/2})\) thus beating transductive matrix completion methods.

\section{Experiments} 
\label{sec:experiments}
  In this section we present an experimental comparison of the proposed sparse-group IMC algorithm (SGIMC) to IMC~\cite{yuetal2014}, as well as the standard matrix factorization (MF) approach to the matrix completion based on stochastic gradient optimization, which serve as baseline methods. We note that we tried to compare SGIMC with another sparse IMC algorithm~\cite{Lu2016} but it run out of memory even on the smallest problems, which we consider. We also compare the algorithms on semi-supervised clustering and matrix completion tasks based on real-world datasets.

\subsection{Synthetic datasets}
\label{subsec:synth}
  In the experiments with synthetic data we consider the regression IMC problem for an \(\nsize_1 \times \nsize_2\) matrix \(\targetMatrix\) for \(\nsize_1 = 800\), \(\nsize_2 = 1600\). We compare different matrix completion algorithms by measuring their relative reconstruction error, given by \(\| \hat{\targetMatrix} - \targetMatrix \|_F / \| \targetMatrix \|_F\), where \(\hat{\targetMatrix} = X \hat{\xFactor} \hat{\yFactor}^{\T} Y^{\T}\). The aim of the study is to show that the feature selection achieved by our algorithm improves the overall generalization ability, while being also competitive in the case with no redundant features.

  The experiment setup is as follows: we generate \(\targetMatrix = X \xFactor \yFactor^{\T} Y^{\T} + \varepsilon\), where the noise matrix \(\varepsilon\) is drawn elementwise from \(\mathcal{N}(0, 0.005)\), and the side-channel features \(X\) and \(Y\) are random Gaussian matrices from \(\mathcal{N}(0, 0.05)\) with dimensions \(\nsize_1 \t d\) and \(\nsize_2 \t d\), respectively. The true values of \(\xFactor\) and \(\yFactor\) are fixed to the first \(k\) columns of the \(d \times d\) identity matrix, thereby making \(k\) the number of informative features.

  We measure the performance as described above with respect to different sparsity levels of \(\targetMatrix\), i.e. the fraction of the observed values \(\rho = \tfrac{\lvert \observedIndices\rvert}{\nsize_1 \cdot \nsize_2}\), and the number of noisy features \(d-k \geq 0\). For the sparsity experiment we sample \(\rho\) percent of entries \(\observedIndices\) from the noisy one to train the models. In each experiment we fix the true rank \(k\) of \(M\) to \(25\) and apply the algorithms with assumed rank \(\hat{k} \in \{20,30\}\), i.e. underestimating and overestimating the true rank. The performance is reported for the lowest error value over the validation subsample of the regularization parameter \(\lambda = \lambda_{\xFactor{}} = \lambda_{\yFactor{}}\) chosen from the set \(\{ {10}^{-5}, {10}^{-4}, {10}^{-3}, {10}^{-2} \}\).

\paragraph{Sparsity \(\rho\).} 
\label{par:sparsity_rho}
  Under this setup both IMC and SGIMC algorithms seem to achieve almost a perfect recovery when matrix \(M\) is relatively dense (\(\rho > 0.1\)), therefore we report the performance in the relatively sparse regime. We set number of features \(d\) to \(100\), vary \(\rho\) from \(0.0005\) to \(0.02\) with the step \(0.0015\).

\begin{figure*}[t!] 
    \centering
    {
    \subfigure[Increasing \(\rho\)]{\label{figsynthnelem}
      \includegraphics[width=0.45\linewidth]{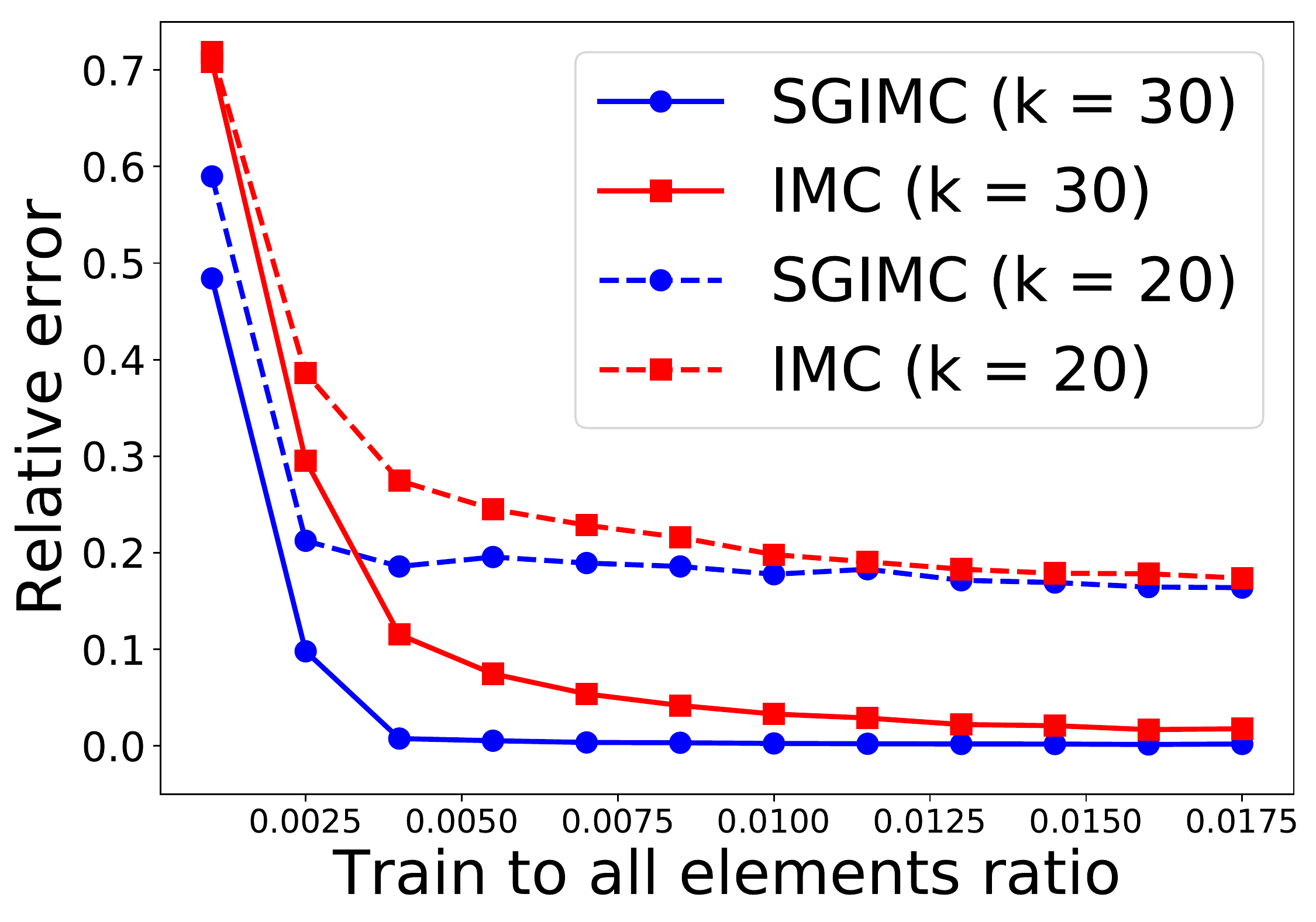}}
    \qquad
    \subfigure[Increasing \(d\)]{\label{figsynthnfeatures}
      \includegraphics[width=0.45\linewidth]{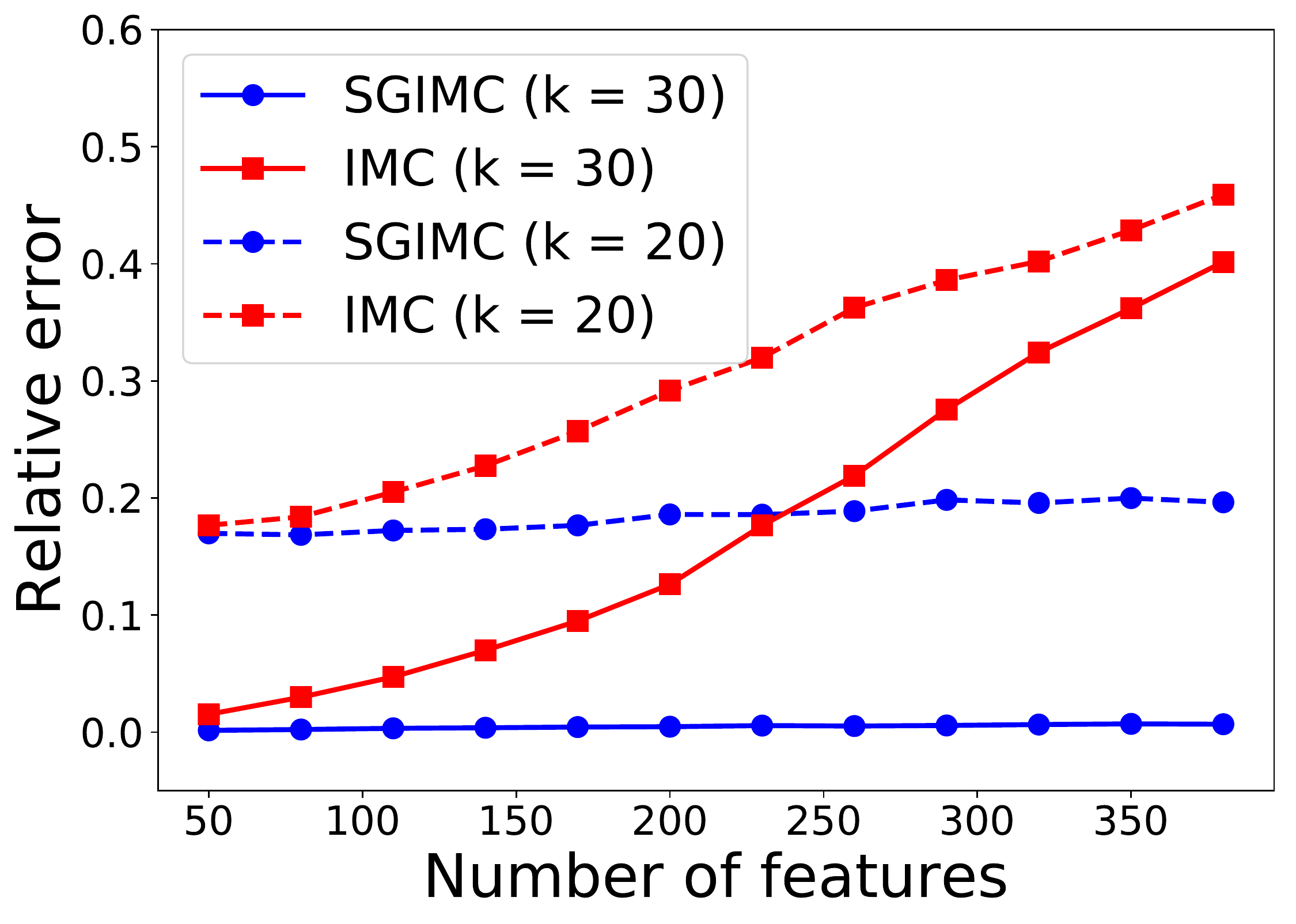}}
    }
    \caption{Experiment on synthetic data: increasing sparsity parameter \(\rho\) (\ref{figsynthnelem}) and increasing number of redundant features \(d\) (\ref{figsynthnfeatures}), respectively.}
  \label{fig:synth}
  \end{figure*}

  The results presented in Figure~\ref{figsynthnelem} suggest that the SGIMC requires fewer observed elements in \(M\) than IMC to achieve a comparable level of reconstruction precision, while overestimation of the rank still allows for solution with zero error.


\paragraph{Redundant features.} 
\label{par:noisy_features}

  This experiment aims at studying a feature selection in the presence of redundant features, by changing the number of features \(d\) from \(50\) to \(400\) in steps of \(50\). The added features are random and not correlated with the target variables. We test the performance for sparsity \(\rho=0.2\).

  Figure~\ref{figsynthnfeatures} indicates that the SGIMC indeed adequately distinguishes informative features from the noisy ones and performs consistently well under both rank regimes.


\subsection{Real-world applications} 
\label{sub:real_world_applications}
  In this section we apply the IMC, the SGIMC, and the MF algorithms to the real-world datasets in order to compare their performances.


\paragraph{Semi-supervised clustering.}
  Given \(\nsize\) items with \(\nsize \t d\) feature matrix \(X\), we consider the problem of predicting if two objects \(i\) and \(j\) belong to the same class. The binary label data is thus a signed similarity matrix \(\targetMatrix\) with \(\targetMatrix_{ij} = 1\) if items \(i\) and \(j\) are from the same class or \(-1\) otherwise.

  Three datasets are chosen to test the matrix completion for the solution of the semi-supervised clustering: Mushrooms, Segment and Covtype~\cite{chang2011libsvm}, see Table~\ref{tb:datasets}. We note that the Covtype dataset is preprocessed to correct for the class imbalance using undersampling based on the size of the most underrepresented class.

  Since we have an access to the full matrix \(M\), we randomly split each dataset into train and test sets with the share of the train sample varying between \(0.0005\) and \(0.02\). The classification accuracy on the test set is used to measure the performance of the algorithms.
  \begin{table}[ht!]
    \centering
    \caption{General information about semi-supervised clustering datasets.}
    \label{tb:datasets}
    \begin{tabular}{lrrr}
      \toprule
            & Mushrooms  &  Segment    & Covtype   \\
      \midrule
      Observations \(n\)    & \(8124\)   &  \(2319\)   & \(7370\)  \\
      Features \(d\)        & \(112\)    &  \(18\)     & \(54\)    \\
      Classes \(K\)         & \(2\)      &  \(7\)      & \(7\)     \\  
      \bottomrule
    \end{tabular}
  \end{table}

  Experiments with the algorithms show that the semi-supervised clustering matrices of each dataset are low-rank and that there are insignificant differences in performance with \(\hat{k}= 2,\ldots, 20\).
  The strategy of choosing regularization parameters and averaging across different random trials was similar to the synthetic experiments, see Section~\ref{subsec:synth}.

  \begin{figure*}[ht!]
    {\caption{Recovery of various semi-supervised clustering datasets.}}
    {
    \subfigure[Mushrooms dataset.]{\label{figmscmush}
      \includegraphics[width=0.3\linewidth]{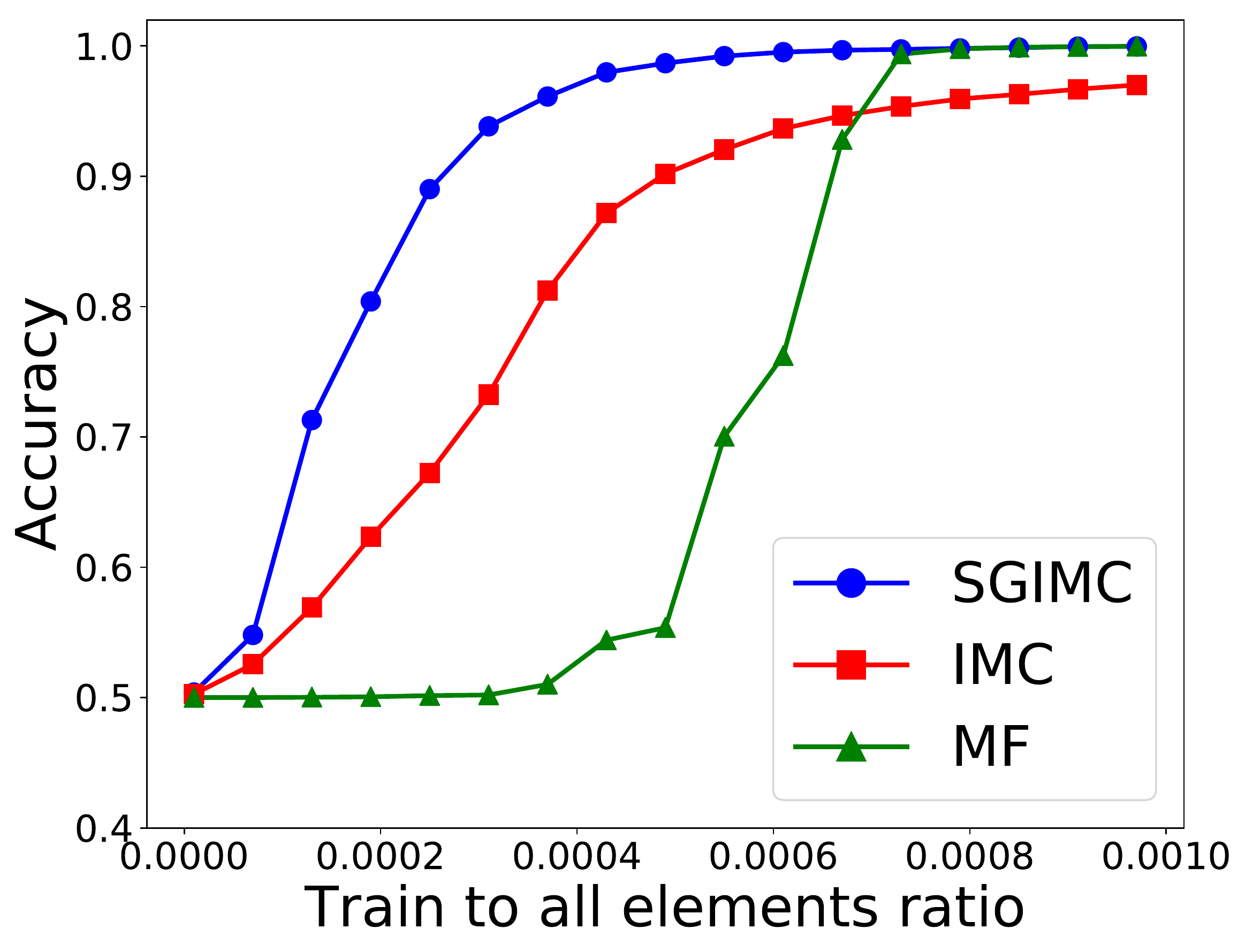}} \hfill
    \subfigure[Segment dataset.]{\label{figmscsegm}
      \includegraphics[width=0.3\linewidth]{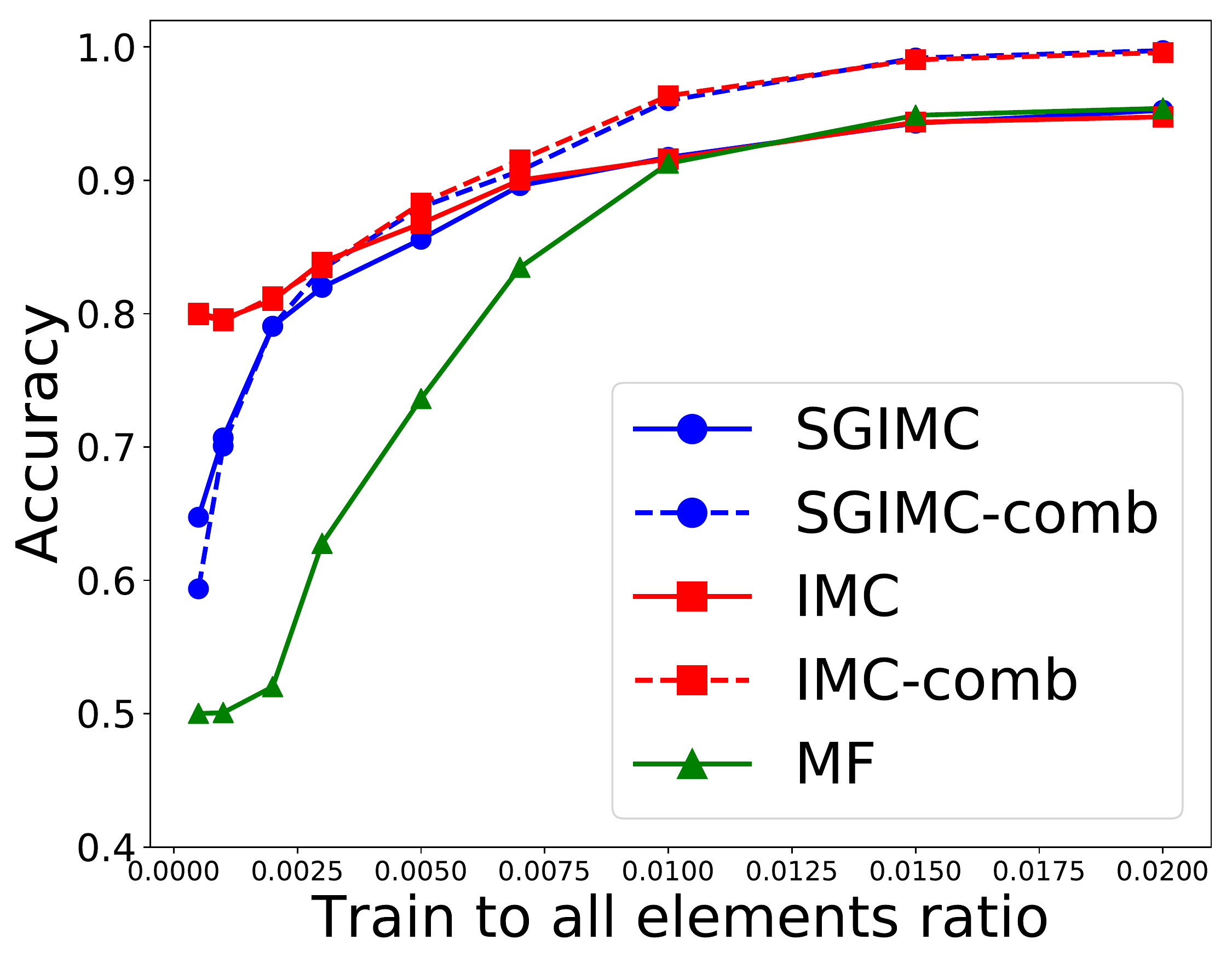}}
      \hfill
    \subfigure[Covtype dataset.]{\label{figmsccovt}
      \includegraphics[width=0.3\linewidth]{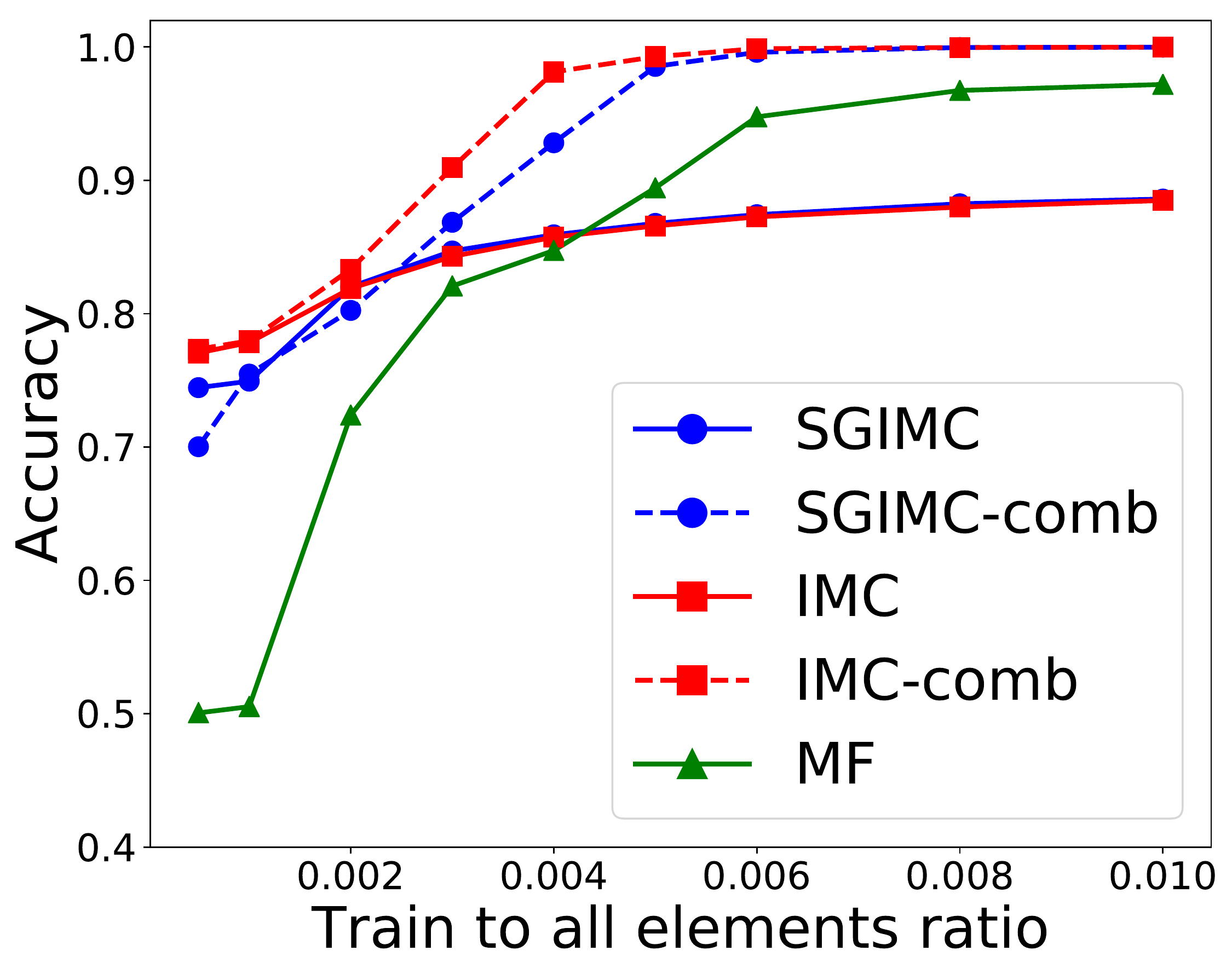}}
    }
  \end{figure*}

  Preliminary analysis shows, that on the Covtype dataset the original side-channel features are insufficient for IMC or SGIMC to achieve the accuracy higher than \(0.9\), even in the overestimated rank regime, Figure~\ref{figmsccovt}. In order to be able to compare IMC and SGIMC results with the ones for MF, we augmented the side-channel features appending an identity matrix to \(X\) from the right. This effectively extended our algorithm to the transductive matrix completion setting similar to~\cite{Chiang2015}. In the results of the experiments (Figures~\ref{figmscsegm} and~\ref{figmsccovt}) the augmented side-channel features are labelled by \emph{SGIMC-comb} and \emph{IMC-comb}.

  \begin{table}[t]
    \centering
    \caption{Accuracy for ``Segment''}
    \label{tb:features-segm}
    \begin{tabular}{lrrr}
      \hline
      Additional features:   & \(0\)           & \(50\)          & \(100\)         \\ \hline
      \multirow{2}{*}{SGIMC} & \(0.901\)       & \(0.885\)       & \(0.880\)       \\
                             & (\(\pm 0.003\)) & (\(\pm 0.003\)) & (\(\pm 0.006\)) \\ \hline
      \multirow{2}{*}{IMC}   & \(0.895\)       & \(0.839\)       & \(0.822\)       \\
                             & (\(\pm 0.007\)) & (\(\pm 0.011\)) & (\(\pm 0.006\)) \\ \hline
      Additional features:   & \(200\)         & \(300\)         & \(400\)         \\ \hline
      \multirow{2}{*}{SGIMC} & \(0.869\)       & \(0.871\)       & \(0.851\)       \\
                             & (\(\pm 0.007\)) & (\(\pm 0.019\)) & (\(\pm 0.014\)) \\ \hline
      \multirow{2}{*}{IMC}   & \(0.795\)       & \(0.769\)       & \(0.754\)       \\
                             & (\(\pm 0.005\)) & (\(\pm 0.006\)) & (\(\pm 0.007\)) \\ \hline
    \end{tabular}
  \end{table}
  %

  %

  
  In Table~\ref{tb:features-segm}  we report the results of a semi-synthetic experiment: we take the real dataset Segment with low feature dimension and introduce non-informative features to the side-channel matrices. The accuracy result demonstrates that feature selection, specifically elimination of non-informative features, is really crucial for the accuracy: the IMC algorithm~\cite{yuetal2014} performs relatively poorly as compared to the proposed approach.

\paragraph{Mycobacterium tuberculosis drug resistance and susceptibility.}
  As it was mentioned, SGIMC helps to filter out efficiently irrelevant side features. It can be very useful when interpretability of the model is essential. The following dataset contains the reactions of \(\numprint{4734}\) \textit{M tuberculosis} strains to \(13\) existing anti-tuberculosis drugs. The task is to predict strain reaction (resistant or susceptible) on each of \(13\) drugs.
  
   Each strain has \(\numprint{355709}\) binary features, indicating the presence of a specific gene mutation, each drug has \(28\) binary features, corresponding to certain chemical properties. The dataset is a combined set of data used in articles \cite{Coll,Walker,Pankhurst,Farhat}.
  %
  %
  Since the number of susceptible instances significantly exceeds the number of resistant instances, we have estimated the prediction quality with \(F_1\) score. \(F_1\) scores for each antibiotic and for all of the together averaged over ten \(50/50\) train/test splits are reported in Table~\ref{tb:tuberculosis-coldstart}. 

  \begin{table}[t]
    \centering
    \caption{\textit{M tuberculosis} \(F_1\)}
    \label{tb:tuberculosis-coldstart}
    \begin{tabular}{lll}
      \hline
      Drugs         & SGIMC         & IMC           \\ \hline
      Isoniazid     & \textbf{0.89} & 0.86          \\ \hline
      Ethambutol    & \textbf{0.62} & 0.61          \\ \hline
      Rifampicin    & \textbf{0.89} & 0.88          \\ \hline
      Pyrazinamide  & 0.53          & 0.53          \\ \hline
      Streptomycin  & 0.84          & \textbf{0.85} \\ \hline
      Ofloxacin     & \textbf{0.48} & 0.42          \\ \hline
      Capreomycin   & \textbf{0.34} & 0.28          \\ \hline
      Amikacin      & \textbf{0.47} & 0.42          \\ \hline
      Moxifloxacin  & \textbf{0.45} & 0.38          \\ \hline
      Kanamycin     & 0.4           & 0.4           \\ \hline
      Prothionamide & 0.52          & 0.52          \\ \hline
      Ciprofloxacin & 0.52          & \textbf{0.67} \\ \hline
      Ethionamide   & \textbf{0.5}  & 0.47          \\ \hline \hline
      Overall       & \textbf{0.59} & 0.57          \\ \hline
    \end{tabular}
  \end{table}
  We conclude that SGIMC not only outperforms IMC on the majority of considered subproblems, but it also achieves a good quality using only about \(\numprint{6000}\) features, while IMC use all \(\numprint{355709}\) features and thus cannot help to explore genetic mechanisms of drug resistance of tuberculosis. 

\section{Conclusions}
\label{sec:conclusion}
  In this work we propose a new inductive matrix completion algorithm which utilizes sparsity inducing penalty to achieve the selection of side-channel features. The method achieves the state-of-the-art performance on both synthetic and real-world datasets; moreover, it outperforms the competitors if many redundant features are present in the data. Our theoretical analysis has shown that the group-sparsity penalty effectively helps to improve generalization bounds and to decrease the sample complexity in sparse problems. We also consider the case when imperfect features are present and provide an extension of the algorithm, which still appears to be consistent when features are not fully predictive.
  
  Our future work is going to deal 
  with designing an algorithm which would be able to find the sparsity patterns that are different from a simple row- and column-wise sparsity. More general sparsity patterns are quite natural for some applications, in particular, for multi-label learning. On theoretical side, there are two presumable directions for improvement. The first one is to design an optimization algorithm which would achieve a global convergence similarly to~\cite{Zhang2018}, but would work with the sparsity inducing penalty and possibly non-quadratic loss function. The second direction is to generalize results of~\cite{Baldin2018} to general inductive matrix completion setting and to obtain a general minimax error bounds for the considered problem.

\subsubsection*{Acknowledgements}
  The reported study was funded by RFBR according to the research project 18-37-00489.

\bibliographystyle{plain}
\bibliography{references}

\clearpage

\appendix

\section{ADMM Intro}
  The Alternating Directions Method of Multipliers (ADMM) has been actively used for solving convex problems of the form
  \begin{equation}
  \label{eq:gen_cvx}
    \begin{aligned}
      & \underset{x \in \RR^d}{\text{minimize}}
      & & f(x) + g(Ax)
      \,,
    \end{aligned}
  \end{equation}
  for \(A\colon \RR^d \to \RR^m\) linear operator and closed proper convex functions \(f\colon \RR^d \to \RR\) and \(g\colon \RR^m \to \RR\), see~\cite{parikhboyd2014}. It has been applied to distributed machine learning problems in~\cite{forero2010}, decentralized optimization~\cite{lingribero2014} and~\cite{bianchihachem2014}, sparse regression models and robust signal reconstruction in~\cite{bioucasdiasfigueiredo2010} and~\cite{fadilistarck2009}. A comprehensive survey and review of the approach can be found in~\cite{boydetal2011,parikhboyd2014} and~\cite{polsonetal2015}.

  Roughly, the idea of the ADMM is to consider an equivalent problem for~\eqref{eq:gen_cvx}, formulate its Lagrangian and solve iteratively for its stationary point~\cite{boydetal2011}. The equivalent problem is obtained by introducing a regularizing quadratic term and redundant linear constraint to~\eqref{eq:gen_cvx}:
  \begin{equation}
  \label{eq:gen_cvx_aug}
    \begin{aligned}
      & \underset{x, z}{\text{minimize}}
      & & f(x) + g(z) + \tfrac1{2\eta}\|Ax - z\|_2^2
      \,, \\
      & \text{subject to}
      & & Ax - z = 0
      \,,
    \end{aligned}
  \end{equation}
  for some fixed \(\eta > 0\). If \(\phi\) is the dual variable for the constraint, then the Lagrangian of~\eqref{eq:gen_cvx_aug} is
  \begin{align}
  \label{eq:gen_cvx_aug_lagr}
    &L(x, z, \phi)
    = f(x) + g(z) + \tfrac1{2\eta}\|Ax - z\|_2^2
    + \tfrac1\eta \langle \phi, Ax - z \rangle
    \notag \\
    &= f(x) + g(z) + \tfrac1{2\eta}\|Ax - z + \phi\|_2^2
    - \tfrac1{2\eta} \|\phi\|_2^2
    \,.
  \end{align}
  By taking the nonlinear Gauss-Seidel iterations w.r.t. the primal variables \((x, z)\), and then performing implicit gradient ascent over \(\phi\), one ultimately gets the scaled ADMM iterations for~\eqref{eq:gen_cvx_aug_lagr}:
  \begin{align*}
    x_{t+1}
    &= \arg \min_{x} f(x) + \tfrac1{2\eta}\|A x - (z_t - \phi_t)\|_2^2
    \,,\\
    z_{t+1}
    &= \arg \min_{z} g(z) + \tfrac1{2\eta}\|z - (A x_{t+1} + \phi_t)\|_2^2
    \,, \\
    \phi_{t+1}
    &= \phi_t + (A x_{t+1} - z_{t+1})
    \,.
  \end{align*}
  Under general conditions these iterations converge to a fixed point, which coincides with the optimal solution of~\eqref{eq:gen_cvx}, see~\cite{bertsekas1989parallel}. With respect to \(x\) and \(z\) the iterations perform proximal steps, which essentially constitute a generalized projection onto the level sets the convex function of \(\eta f\) and \(\eta g\), respectively~\cite{parikhboyd2014}.

\section{Tools for theoretical analysis}
  Here and below we assume that \(p\) is a probability distribution on a set \(\zSet \t \CC\), where \(\zSet\) is a feature space and \(\CC\) is a space of outputs (class labels or regression values). Suppose that \(D = \bigl((Z_1, C_1), \dots, (Z_{\msize}, C_{\msize})\bigr)\) are i.i.d. samples from \(p\). Consider an estimation function \(f\colon \zSet \to \CC\) from some functional class \(\FC\). If the loss function \(\LC\colon \CC \t \CC \to \RR_+\) is given then the main question of statistical learning is whether we can choose function \(f \in \FC\) based on dataset \(D\) in such a way, that on average for random pair \((Z, C)\) from distribution \(p\) loss \(\LC(f(Z), C)\) is small. Here and below for the ease of notation we assume that loss values for the considered problems are bounded and belong to the interval \([0, 1]\). 
  
  We start by reminding the notion of Rademacher complexity, which is used as the basic argument in the analysis.

  \begin{definition}
    Let \(\FC\) be a function class mapping from \(\zSet\) to \(\RR\). Then the empirical Rademacher complexity is defined by
    \begin{EQA}[c]
    \label{eq: rademacher}
      \widehat{ \RC}_{\msize}(\FC) = \EE \Biggl[\sup_{f \in \FC} \sum_{i = 1}^{\msize} \eps_i f(Z_i) \biggl| Z_1, \dots, Z_{\msize}\Biggr].
    \end{EQA}
  \end{definition}

  The key lemma from~\cite{Maurer2012} allows to bound the expected risk of wrong prediction by empirical risk and Rademacher complexity of considered functional class (see also~\cite{Bartlett2002}). 
  \begin{lemma}
    Consider a loss function \(\LC\colon \RR \t \RR \to [0, 1]\) with Lipshitz constant \(\lipshitzConst\). Let \(\FC\) be a function class mapping from \(\zSet\) to \(\RR\) and let \((Z_i, C_i)_{i = 1}^{\msize}\) be an i.i.d. sample from measure \(p\). Then for any integer \(\msize\) and any \(\delta \in (0, 1)\) with probability at least \(1 - \delta\) (over the samples of length \(\msize\)) and for every \(f \in \FC\) it holds
    \begin{EQA}
      \EE \LC(C, f(Z))
      &\le&
      \frac{1}{\msize} \sum_{i = 1}^{\msize} \LC(C_i, f(Z_i))
      +
      \lipshitzConst \widehat{\RC}_{\msize}(\FC)
      \\
      &+&
      \sqrt{\frac{9 \ln 2 / \delta}{2 \msize}}.
    \end{EQA}
  \end{lemma}
  Thus, to bound the risk we need to bound the Rademacher complexity \(\widehat{\RC}_{\msize}(\FC)\). For this purpose we use the following theorem from~\cite{Maurer2012}, which bounds Rademacher complexity for the important special case of linear estimators. Here and below we consider the realizable case \(C = f(Z)\) and with slight abuse of notation denote the distribution of \(Z\) as \(p\). Let us also denote \(\ZV = (Z_1, \dots, Z_{\msize})\).
  \begin{theorem}
  \label{theorem:rademacherlasso}
    Let \(\zSet = \RR^{d}\), \(\ZV = (Z_1, \dots, Z_{\msize})\) be an i.i.d. sample from \(p\) and \(\FC = \{f\colon \RR^{d} \to \RR ~ s.t. ~ f(Z) = \langle \beta, Z \rangle, ~ \|\beta\|_1 \le 1\}\). Let \(\widehat{\RC}_{\msize}(\FC)\) be defined as in~\eqref{eq: rademacher}. Then
    \begin{EQA}[c]
    \label{lasso_bound}
      \widehat{\RC}_{\msize}(\FC) \le \frac{2^{3/2}}{\msize} \sqrt{\sum_{i = 1}^{\msize} \|Z_i\|_{\infty}^2} (2 + \sqrt{\ln d}).
    \end{EQA}
  \end{theorem}
  The stronger variant of this result can be also found in~\cite{Maurer2012}. 
  \begin{theorem}  
  \label{theorem:rademacher_expectation}
    Let \(\PC\) be a set \(\left \{ P_1, P_2, \dots, P_{d} \right\}\) of orthogonal projections \(P_k\) onto the 1-dimensional subspace generated by the basis vector \(e_k\). Then under the conditions of Theorem~\ref{theorem:rademacherlasso} we have
    \begin{EQA}
      && \widehat{\RC}_{\msize}(\FC)
      \le
      \frac{2^{3/2}}{\msize}\sqrt{\sum_{i = 1}^{\msize} \|Z_i\|_{\infty}^2}
      \\
      &\cdot& \left (2 + \sqrt{\ln\frac{1}{\msize}\sum_{i = 1}^{\msize} \sum_{P \in \PC} \left|P Z_i \right|} \right)
      +
      \frac{2}{\sqrt{\msize}}.
    \label{accurate_lasso_bound}
    \end{EQA}
  \end{theorem}
  The bound~\eqref{accurate_lasso_bound} allows, in particular, to obtain tighter upper bound compared to~\eqref{lasso_bound} in case when features \(Z_i\) have sparse coordinates. We illustrate this case below.
  
\subsection{Proof of Theorem~\ref{theorem:main}}
\label{sec:proofTheorem}
  We note that the considered bounds on parameters in~\eqref{sparse_imc_model_theory} allow for the solution a giving zero empirical error \(\hat{r}_{\msize}(\xFactor, \yFactor)\). Thus, in order to assess the generalization ability of the considered algorithm and to bound the expected risk \(r_{\msize}(\xFactor, \yFactor)\) we need to bound empirical Rademacher complexity \(\widehat{\RC}_{\msize}(\FC)\).
  
  We start by the following representation of the regression function:
  \begin{EQA}
    && \sup_{\|\xFactor\|_{2, 1} \le \xFactor_{2, 1}, \|\yFactor\|_{2, 1} \le \yFactor_{2, 1}} \langle \xFactor^{\T} \xv, \yFactor^{\T} \yv \rangle
    \\
    &=&
    \sup_{\|\xFactor\|_{2, 1} \le \xFactor_{2, 1}, \|\yFactor\|_{2, 1} \le \yFactor_{2, 1}} \langle \xFactor \yFactor^{\T}, \xv \yv^{\T} \rangle
    \\
    &=&
    \sup_{\parameterMatrix = \xFactor \yFactor^{\T}, \|\xFactor\|_{2, 1} \le \xFactor_{2, 1}, \|\yFactor\|_{2, 1} \le \yFactor_{2, 1}} \langle \parameterMatrix, \xv \yv^{\T} \rangle.
  \end{EQA}
  Thus, we are working in the framework of linear models and can use the result of Theorem~\ref{theorem:rademacherlasso}. The main question is which norm we can bound for matrix \(\parameterMatrix\). The essential choice is the bound for \(\|\parameterMatrix\|_{1, 1}\):
  \begin{EQA}
    \|\parameterMatrix\|_{1, 1}
    &=&
    \sum_{i = 1}^{d_1} \sum_{j = 1}^{d_2} |\uv_i^{\T} \vv_j|
    \le
    \sum_{i = 1}^{d_1} \sum_{j = 1}^{d_2} \|\uv_i\|_{2} \|\vv_j\|_{2}
    \\
    &=&
    \|\xFactor\|_{2, 1} \|\yFactor\|_{2, 1}.
  \end{EQA}
  This bound allows to directly apply Theorem~\ref{theorem:rademacherlasso} and get
  \begin{EQA}
    \widehat{\RC}_{\msize}(\FC)
    &\le&
    \frac{2^{3/2}}{\sqrt{\msize}} \xFactor_{2, 1} \yFactor_{2, 1} \bigl(2 + \sqrt{\ln (d_1 d_2)}\bigr)
    \\
    &\le&
    s_1 s_2 k u_{\infty} v_{\infty} \frac{2^{3/2}}{\sqrt{\msize}} \bigl(2 + \sqrt{\ln (d_1 d_2)}\bigr).
  \end{EQA}
  Note that in the case when the design vectors are sparse we can improve the bound by using Theorem~\ref{theorem:rademacher_expectation}. Let us assume that for some vectors \(\xv\) and \(\yv\) it holds that
  \begin{EQA}
  \label{sparsity_cond}
    &&\sum_{i=1}^{d_1} I\left\{\xv^{(i)} \neq 0 \right\} = r_1 \le d_1,
    \\
    &&\sum_{j=1}^{d_2} I\left\{\yv^{(j)} \neq 0 \right\} = r_2 \le d_2.
  \end{EQA}
  Then we can obtain for the projections onto the 1-dimensional subspaces that
  \begin{EQA}[c]
    \sum_{i=1}^{d_1} \sum_{j=1}^{d_2} \left|\xv^{(i)} \yv^{(j)}\right| \leq r_1 r_2.
  \end{EQA} 
  Thus, if all the feature instances \(\xv_i, \yv_j\) satisfy the conditions~\eqref{sparsity_cond} then by Theorem~\ref{theorem:rademacher_expectation} we obtain
  \begin{EQA}[c]
    \widehat{\RC}_{\msize}(\FC) \le s_1 s_2 k u_{\infty} v_{\infty} \frac{2^{3/2}}{\sqrt{\msize}} \left(2 + \sqrt{\ln (r_1 r_2)}\right) + \frac{2}{\sqrt{\msize}}.
  \end{EQA}
\end{document}